\documentclass[preprint,12pt]{elsarticle}
\usepackage{float}
\usepackage{etoolbox}
\usepackage[noend]{algpseudocode}
\usepackage{algorithm}
\usepackage{algorithmicx}

\usepackage{graphicx}
\usepackage{amssymb}
\usepackage{amsmath}

\usepackage{lineno}
\usepackage[normalem]{ulem}
\usepackage{color}
\journal{Talanta}

\begin{document}
	
	\begin{frontmatter}
		
		
		\title{Uncharted Forest: a Technique for Exploratory Data Analysis}
		

		
		\author{Casey Kneale}
		\author{Steven D. Brown\corref{cor1}}
		\cortext[cor1]{sdb@udel.edu}
		\address{Department of Chemistry and Biochemistry \\ 163 The Green, University of Delaware, Newark, Delaware, 19716, USA}
		
		\begin{abstract}
			Exploratory data analysis is crucial for developing and understanding classification models from high-dimensional datasets. We explore the utility of a new unsupervised tree ensemble called uncharted forest for visualizing class associations, sample-sample associations, class heterogeneity, and uninformative classes for provenance studies. The uncharted forest algorithm can be used to partition data using random selections of variables and metrics based on statistical spread. After each tree is grown, a tally of the samples that arrive at every terminal node is maintained. Those tallies are stored in single sample association matrix and a likelihood measure for each sample being partitioned with one another can be made. That matrix may be readily viewed as a heat map, and the probabilities can be quantified via new metrics that account for class or cluster membership. We display the advantages and limitations of using this technique by applying it to two classification datasets and three provenance study datasets. Two of the metrics presented in this paper are also compared with widely used metrics from two algorithms that have variance-based clustering mechanisms.
		\end{abstract}
		
		
	\end{frontmatter}

	
	\section{Introduction}
	\label{S:11}
	
	Chemometric classification methods are often used to discriminate high-dimensional chemical signatures of unknown samples to determine their most likely class label \cite{patternrec}. These methods have proven to be valuable for the fields of archeaometry and forensics where the origin, or provenance, of a manufactured item is often of interest\cite{patternrec, obsidian, crypto, shipwreck, butter, heroin} because the chemical data, especially those obtained from neutron activation analysis (NAA) \cite{sayre,NAArev}, tend to be highly multivariate. The high dimensionality of data poses a challenge for understanding sample relationships because it cannot be easily visualized or interpreted directly. While, these methods can succeed at classifying samples, they do so by including information from the class labels into the model, and they provide little insight about trends or patterns found in the data with the absence of labels. Information pertaining to which samples are most similar to one another between classes, or which are different from the rest of the samples within a given source class, without the influence of label discrimination is not available from classification methods. 
	
	Exploratory methods are often used to reveal trends and other patterns hidden in data without the use of class labels \cite{patternrec}. Exploratory data analysis (EDA) is a form of data analysis which encompasses a number of visual exploration methods. Success in an EDA study depends on the creativity of the analyst as much as on the technique. Although there is no strict definition of EDA, it has been stated that, ``Exploratory data analysis isolates patterns and features of the data and reveals these forcefully to the analyst" \cite{tukeybook}, and also that ``Exploratory data analysis’ is an attitude, a state of flexibility, a willingness to look for those things that we believe are not there, as well as those we believe to be there"\cite{tukeybooktwo}. The coupling of EDA methods with preprocessing and/or feature selection can aid an analyst in the discovery of patterns, or the lack thereof, in data. The information gained from EDA may be used with feature selection, preprocessing and modeling methods to iteratively improve a data analysis pipeline.
	
	The two predominant types of EDA focus on dimension reduction and clustering. These methods have been extensively used for the EDA of archeometric data \cite{crypto, shipwreck,  butter, cluster} because high dimensional data are not readily visualized. Dimension reduction techniques such as principal component analysis (PCA), discriminant analysis, or exploratory projection pursuit are used to change the basis of the data to one based on lower-dimensional projections \cite{patternrec}. Unfortunately, these methods do not offer a guarantee that the resulting 2-D or 3-D projections will offer meaningful information about class relationships or sample-class relationships, due to the presence of class overlap or class heterogeneity\cite{patternrec}. Unsupervised classification as implemented in clustering algorithms suffers from the opposite problem clustering methods readily create groupings of high-dimensional data but tend to offer little information about the cluster assignments themselves. Clustering methods have been known to incorrectly associate archeological samples depending on the method used and on the chemistry of the samples \cite{cluster}. Information about sample or class relationships may be inferred by employing many techniques, but the resulting information is often hidden, and typically key relationships in the data cannot be seen by an analyst \cite{cluster}.
	
	This work reports a new approach, which we call the uncharted forest, to the visualization and measuring of relationships within and between classes of data. This approach has elements of a clustering algorithm in that it groups similar samples with one another, but is also similar to dimension reduction methods in that it outputs a single heat map which can be interpreted to reveal information about the samples. Uncharted forest analysis uses a partitioning method that is related to the sample partitioning approach used in decision trees but, it does not use class labels like most tree methods do \cite{CARTBook}. Instead, the uncharted forest analysis explores how samples relate to one another under the context of univariate variance partitions. Although the method is unsupervised, we show that when the results are overlaid with external class labels, the method can be used to investigate sample relationships as they pertain to class labels. We demonstrate that this technique can be used as a tool for exploratory data analysis to visualize class or cluster associations, sample-sample associations, class heterogeneity, and uninformative classes. The utility of uncharted forest analysis is demonstrated on two classification datasets and two provenance datasets. Additionally, two empirical clustering metrics are compared with two of the metrics obtained by uncharted forest analysis on another provenance dataset.

	\section{Theory}
	To motivate the use of an unsupervised tree ensemble, a brief review of supervised trees and bagged tree ensembles as used in classification is provided here. A complete review of these methods can be found elsewhere \cite{CARTBook,BreimanRF}. 
	
	\subsection{Supervised Classification Trees}
	Before describing the mechanisms by which supervised decision trees are used to partition data, a survey of the vocabulary common to these methods is presented. Classification trees are a type of supervised classifier, which means that the assignment of new data to a class label requires that each sample the tree was trained on also has a label. The aim of developing a classification tree is the establishment of a set of sequential rules that can be applied to label a sample based on its features. Supervised decision trees are collections of many binary decisions, where each decision is made at one of three locations: roots, branches, or terminal nodes/leaves. These three locations are displayed in their hierarchical ordering in Figure 1. A root is the first decision made in the tree. Branch nodes indicate later, non-terminal decisions. Terminal nodes indicate where a branch has been terminated and where final decisions that assign class labels are made. 
	
	\begin{figure}[h]
		\centering
		\includegraphics[width=0.40\linewidth]{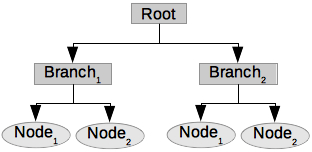}
		\caption{A block diagram showing the relationships between roots, branches and terminal nodes via arrows.}
	\end{figure}
	
	Every sample from the training set is partitioned at a branch or node based on whether a selected variable for the sample has a value that is greater than or less than a specified value. The combination of a variable and value that partitions a set of samples is often referred to as a decision boundary. Decision boundaries are obtained by exhaustively searching each variable and finding the threshold value which affords the highest gain \cite{CARTBook}. Here, gain is a metric-dependent measure of how well a decision separates the available samples according to their class labels. Metrics that can be used to minimize class label impurities at terminal nodes, such as the Gini impurity, informational entropy, and classification accuracy are commonly used to train supervised classification trees by finding a set of decision boundaries \cite{CARTBook}. 
	
	For example, if the samples available at a branch node are better separated into their respective classes by application of some decision, two new branches are made at that node. If not, the partitioning terminates, the node becomes a terminal node, and the decision tree ceases growth using those samples. In this way, each observation is sorted from the root of a tree through branches and eventually to terminal nodes, based on whether creating a new decision boundary better separates samples by known class labels\cite{CARTBook}. 	
	
	One advantage to the use of decision trees is that the classifier that results can be easily understood by examining the decision boundaries that partition the data and the relation of these boundaries to class labels. However, the predictive performance of classification trees on new samples is generally poor relative to other models because the partitions are sensitive to noise in the variables \cite{CARTez}. Tree-based classifiers also tend to overfit the data, and some sort of branch removal, or pruning step, is usually employed to reduce the complexity of the tree \cite{CARTBook}. 
	
	\subsection{Random Forest Classification}
	The random forest classifer is a multiclass classification technique that utilizes the results from an ensemble, or collection of supervised decision trees \cite{BreimanRF} to assign class labels. In this technique, a user-specified number of supervised decision trees is created, each overfit to a selection of bootstrapped samples and random selections of variables. Random forests make use of the complexity of many, biased decision trees to make robust classification assignments by using the average class designation assigned by every tree to predict overall class membership for new data. This approach to ensemble model building is referred to as bootstrap aggregating, or bagging \cite{bagging}. The average of many, complex trees tends to perform well in practice \cite{BreimanRF}.
	
	Random forest classification has some distinct advantages relative to many other classifiers. It is robust to outliers because sample selection at branches is bootstrapped, and the effect of a few outlying samples does not affect its cost function significantly \cite{BreimanRF}. Random forest models are also robust to noise in variables because many classifiers, each with randomly selected variables, are used to decide on class memberships. Another notable advantage to random forest classification is that there is no requirement for linearity in class boundaries. Random forest models, unlike methods such as linear discriminant analysis or support vector machines, are not based on the hypothesis that classes are linearly separable \cite{linearmethods,kernel}. Random forest classification also requires relatively few hyperparameters and works without the need for significant tuning of the classifier on many kinds of data \cite{CARTez}. 
	
	\subsection{Uncharted Forest Tree}
	The concept of unsupervised decision tree modeling is relatively new to chemometric applications. Only a few approaches using unsupervised trees have been reported. An approach reported by Khudanpur, et al. is based on an algorithm similar to the one reported here, but their approach uses the Kullback-Leibler distance to find sets of unsupervised partitions that are optimal with respect to an information theoretic measure 
	\cite{InformationApp}. Other methods for developing unsupervised decision trees are based on the use of an assumed class label for each observation; the trees are trained in a manner that is similar to that used for training a random forest classifier \cite{UnsupRF, CLTrees}. 
	
	Our approach to unsupervised decision trees, the uncharted forest tree algorithm, focuses on intuitive concepts from statistics or machine learning rather than on information theory, to allow for easy interpretation. The trees can be created without the need for class labels because the gain function that is optimized in the construction of each tree relies only on information in the data matrix, not on the labels. This algorithm does not utilize the usual supervised metrics for optimization such as the Gini importance or entropy \cite{BreimanRF}. Instead, the tree hierarchies used in uncharted forest are constructed from decision boundaries that reduce measures of spread in a given variable. The reduction of spread is common in pattern recognition and chemometrics; it is the mechanism underlying techniques such as K-means clustering, and principal component analysis for dimension reduction \cite{patternrec}. Metrics based on changes in the variance, median absolute deviation (MAD), or mean absolute deviation (AD) can also be used for creating the decision boundaries used in an uncharted forest tree. If variance is used, however, univariate Gaussian probabilities may be calculated from the samples on either side of a decision boundary. We prefer the variance metric because it is related to several established figures of merit for clustering, as we describe later. The gain metric based on changes in variance that we use is
	
	\begin{equation}
	gain = 1 - \frac{Var(Branch_1) - Var(Branch_2)} {Var(Parent Node)}
	\end{equation}
	
	\noindent where the sample data in either partition are denoted as $Branch_1$ and $Branch_2$, and the sample data prior to introducing the partition is contained in the previous - or parent - node, which has been denoted as, $Parent Node$. The gain in the unsupervised case can then be seen as the loss of variance ($Var$) in the data resulting from the construction of a given decision boundary. The gain in the unsupervised case can then be seen as the loss of variance in the data resulting from the construction of a given decision boundary, relative to that in the previous - or parent - node. Similar gain equations can be written for measures of spread based on AD and MAD. These are not considered in this work because they do not have direct comparisons to clustering metrics. 
	
	The spread in data can almost always be reduced by partitioning data around an available decision boundary. Thus, uncharted forest trees require a user-defined parameter, the tree depth, to limit the number of possible partitioning decisions made. Tree depth can be defined as the maximum number of branches that are counted from the root node to a given terminal node. For exploratory analysis of classification or cluster data, a reasonable depth may be obtained from the rounded base 2 logarithm of either the number of known classes or the number of clusters selected. The number of classes/clusters available is often known information; for example, in most provenance studies the number of classes is decided upon by the number of sampling sites used. In our implementation the minimum node size, or when branch growth is ceased after the number of samples becomes fewer than a given number, has been set to five samples because five samples allows for a calculation of variance that is not undersized but still accommodates a small cluster/class size. Pseudocode that depicts the recursive algorithm for constructing an uncharted decision tree is shown in Algorithm 1.

	
	\begin{algorithm}
		\caption{Uncharted Forest Decision Tree}\label{unchartedtree}
		\begin{algorithmic}[1]
			\Procedure{  }{}
			\State for each variable in variables
			\State \ \ \ \ Calculate gain. (Eqn. 1)
			\State \ \ \ \ Store highest gain.
			\State end for
			\State Partition samples at largest gain.
			\If {branch depth $\geq$ number of branches} 
			\State Stop growth.
			\Else
			\State For each partition \textbf{goto} 2
			\EndIf
			\State \textbf{end if}
			\EndProcedure
		\end{algorithmic}
	\end{algorithm}
	
	The uncharted forest decision tree partitions a set of data by finding decision boundaries that reduce the spread in the data. It functions as a kind of clustering, where the clusters result from divisive partitioning of spread in the data. Unfortunately, unsupervised trees of the form described above often have marginal utility in practice. These trees can be used for clustering but, like a single supervised tree, the individual uncharted forest decision trees are sensitive to noise effects in the data. The variables in a set of data have many contributions to their spreads, and as a consequence, the partitions that result from a single unsupervised tree tend to be uninformative.
	
	\subsection{Uncharted Forest}
	One solution to the uninformative partitions created by a single uncharted forest decision tree is to create an ensemble of unsupervised trees, an ``uncharted forest". The approach is similar to that taken in the random forest classifier. For each tree in this uncharted forest, the variables used are randomly selected, so that potentially relevant information within an individual variable is not obscured by other contributions to its spread or by another variable with much larger gain in the reduction of spread. The algorithm for the uncharted forest method follows that of the random forest algorithm \cite{BreimanRF}, except that the supervised classification and regression tree models are replaced with the uncharted forest decision tree and the samples are not bootstrapped.
	
	Because construction of the trees is unsupervised, the output of the uncharted ensemble algorithm differs from that of the random forest. An uncharted forest returns a square matrix whose dimensionality is that of the number of observations, so that each row or column vector stores information about a single observation and how it relates to the other samples. The matrix is populated in the following manner: when samples are partitioned into a terminal node together, the row and column representative of each of these samples each gain a +1 count to account for their associations with one another. Similarly the row and column of any samples that were not present at a terminal node with those samples each receive a -1 count. The associations formed by random chance become less pronounced by including the -1 penalty on observations that were not grouped with one another at their respective terminal nodes. After row-wise normalization, the matrix possesses diagonal symmetry. In the case of variance-only partitions, a Gaussian probability for node membership may be used instead of $\pm$ 1. 
	
	
	\subsection{Figures of Merit}
	We refer to the matrix generated by the tree ensemble as a probabilistic affinity matrix, because of its resemblance to the distance-based affinity matrices utilized by clustering methods \cite{spectral}. Each element in the matrix represents the row-wise, likelihood that a given row (observation) is associated with a given column (another or the same observation) at a terminal node in an uncharted forest tree. In order to maintain the above interpretation, a ramp function is applied to every matrix element (p), max(0, p), to remove any negative entries. In this way, the diagonal elements have a sampled probability of association of 1.0, because there is a 100\% likelihood that a given sample will be associated with itself at each terminal node where it was assigned. 
	
	We introduce two schemes that allow for the probability values in the matrix to be quantified for exploratory analysis. Two figures of merit can be calculated to measure how often samples between- or within- clusters/classes are partitioned together. The submatrix of the probability affinity matrix (P) that contains the associations between two clusters/classes i and j, can be quantified by what we define as the interaction quotient (IQ),
	
	\begin{equation}
	IQ_{i,j} =  \frac{\sum_{c=j_{min}}^{j_{max}}  \sum_{r=i_{min}}^{i_{max}}   P_{r,c}} {(j_{max} - j_{min}))(i_{max} - i_{min}) }\end{equation}
	
	\noindent where the min and max subscripts denote the contigious row/column beginning and end for cluster/class indices within the matrix. The IQ is simply the sum of all probabilities inside of the submatrix that is defined by the rows shared from cluster/class $i$, and the columns shared by cluster/classes $j$ normalized by the maximum possible probabilities for association within that submatrix. Similarly, we define the self association quotient (SAQ) as a measure of within- class or cluster sample associations. The SAQ metric is calculated from the probabilistic affinity matrix for the special case where $i = j$. For purposes of visualization, the matrix blocks from which these metrics are calculated are displayed below,
	
	\[
	\begin{bmatrix}
	SAQ_{1} & IQ_{1,2} & IQ_{1,3} & \dots  & IQ_{1,j} \\
	IQ_{2,1} & SAQ_{2} & IQ_{2,3}  & \dots  & IQ_{2,j} \\
	\vdots & \vdots & \vdots & \ddots & \vdots \\
	IQ_{i,1} & IQ_{i,2} & IQ_{i,3} & \dots  & SAQ_{i}
	\end{bmatrix}
	\]
	
	The total interaction quotient (TIQ) which describes the extent of the total interactions between the samples of all class/cluster assignments can be expressed by the sum of probabilities within each IQ block divided by the number of matrix entries in those blocks(n),
	
	{\centering
		\begin{equation} TIQ = \sum_{i}  \sum_{j}   IQ_{i,j} / n   \ \ \ \ \ provided \ \ i \neq j \end{equation} \par
	}
	
	In an analogous manner, the total self association quotient (TSAQ) can be used to describe how often the samples in all class/clusters are partitioned with samples in their same respective class/cluster by uncharted forest,
	
	\begin{equation}TSAQ = \sum_{i}   SAQ_{i} / n \end{equation}
	
	The metrics SAQ and IQ provide a means to assess the likelihoods that samples are within the same terminal nodes after variance partitioning with reference to class/cluster ownership. SAQ can provide a measure of class/cluster heterogeneity by indicating how likely it is that the samples inside of a given label are partitioned together. The IQ provides a measure of how often data were associated to the samples that belong to a different class/cluster label. The IQ measure can therefore be used as an aid to identify classes in the data that do not associate with samples from other classes of data, or classes which are similar to one another. These measures can also be made on a sample-wise basis to discover outliers which may either be between class/cluster labels or are not associated with any class/cluster label at all.

	\section{Scope of the Uncharted Forest Method}
	There is one restriction to the use of uncharted forest analysis that should be noted. The associations that a uncharted forest analysis can describe are often not meaningful without data ordered by class or cluster labels, despite that it is an unsupervised technique. The hypothesized membership of samples allow interpretations of the data as it pertains to class/cluster labels. In some cases, data have serial correlations, such as in time series, or variables from spectroscopic instruments, and the probabilistic affinity matrix appears well-ordered. However, with uncorrelated, labelless data, the probabilistic affinity matrix most often appears similar to random noise. One interesting feature of the probabilistic affinity matrix is that it is invariant to the ordering of data. This is a useful attribute when the matrix could display seemingly random disjoint probabilities, because the order of the rows and columns of the matrix can be changed to permit a more meaningful interpretation, like a cluster for example, without running the algorithm again. The studies explored in this work address the case where class or cluster labels were available for provenance and classification data.
	
	\section{Selected Examples}
	In this work, five datasets were investigated by uncharted forest analysis. The first, the well-known iris dataset, is presented as an instructional example to demonstrate how the technique can be used to examine data with class labels, and how it compares to the widely used principal components analysis method. The well-known ARCH obsidian data is presented to compare the sample-sample associations obtained by uncharted forest analysis with two supervised classification methods. The Nan'ao One shipwreck and cryptocrystalline sillicate datasets are presented to compare the results from uncharted forest analysis to those of published works that used other data analysis methods on the same data. The fifth dataset, the butter data, is used to compare the new method to established variance-based partitioning methods and to show that the TIQ and TSAQ figures of merit are similar to well-established, variance-based clustering metrics.
	
	\subsection{Iris Data EDA}
	The iris dataset contains measurements of 150 iris flower samples. Both the widths and lengths of the sepals and petals of each flower were measured. Each sample belongs to either the virginica, versicolor, or setosa species \cite{iris1, iris2}. The iris data is widely regarded as a standard dataset for introducing classification problems and algorithms because it features two elements common to classification problems: it possesses a linearly separable class (setosa) and two classes with overlapping boundaries (virginica, versicolor).
	
	When the uncharted forest algorithm was applied to the iris data using 100 trees, a depth of 4, and using the variance metric, the probabilistic affinity matrix indicated associations between several samples of virginica and versicolor. The row-wise total interaction quotients for the samples in the setosa class were $\ll$ 5\%; this suggested that the setosa class was well separated from the others. Four samples that were greater than 3$\sigma$ from the mean row-wise IQ were found in both the virginica and versicolor classes. This result implies that, by making consecutive decision boundaries to reduce variance, the virginica and versicolor classes had several samples that appeared to share the same class label space. 
	
	Principal components analysis was employed to reduce the dimensions of the iris data and to visually assess which samples uncharted forest analysis indicated as associated with other classes. When the iris data were projected onto the first two principal component axes, five samples which had significant row-wise IQ's were located between the virginica and versicolor classes, as can be seen in Figure 2. A conventional random forest model with 500 trees was then trained on the entire dataset. The random forest model misclassified 5 samples. Of those 5 samples, 3 were the samples with row-wise IQ values that exceeded 3$\sigma$. The heat map from uncharted forest algorithm highlighted many of the samples that the random forest classifier had difficulty classifying due to class overlap.
	
	\begin{figure}[H]
		\centering
		\includegraphics[width=0.9\linewidth]{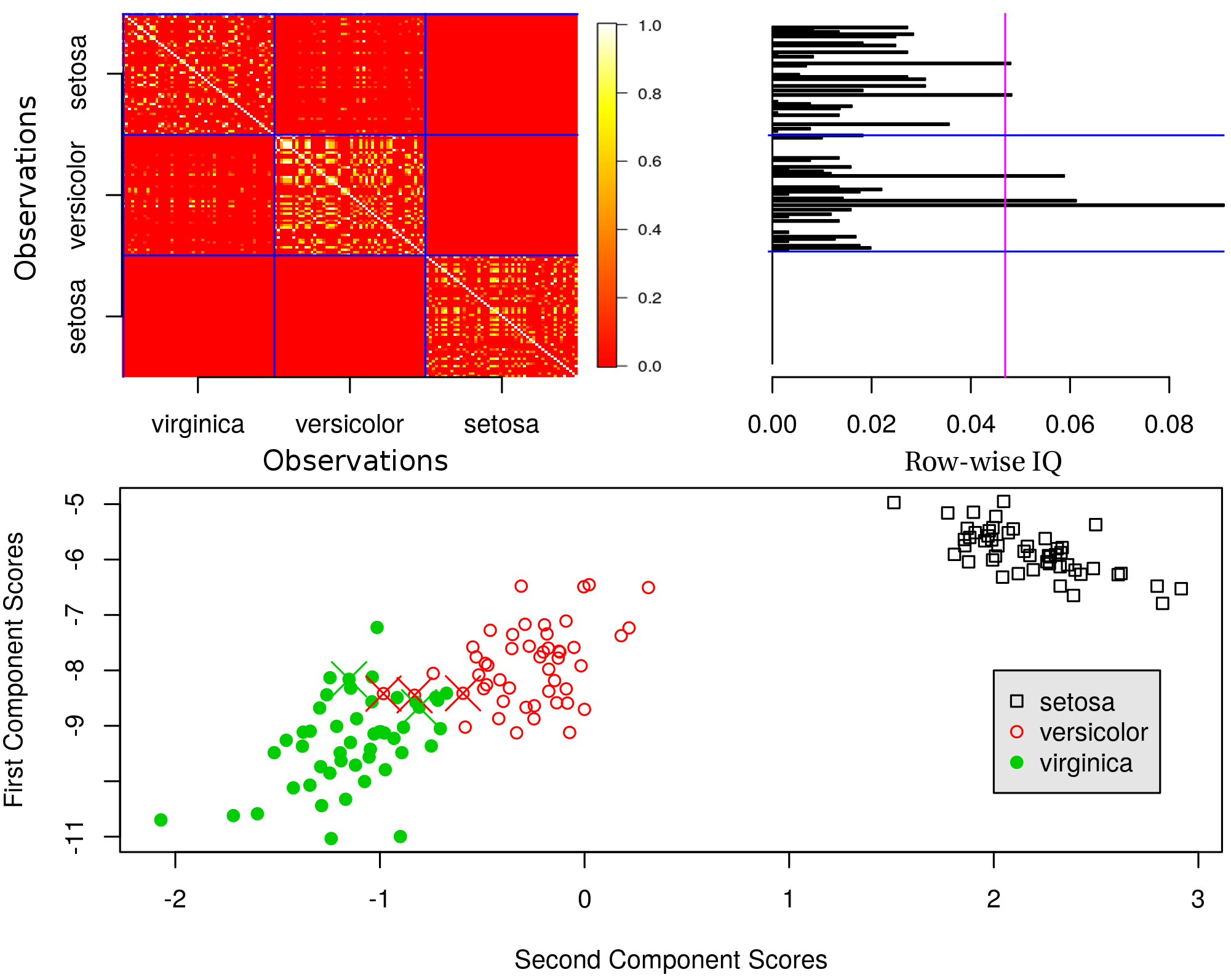}
		\label{fig:Ng1} 
		
		\caption{A heat map of the probabilistic affinity matrix of the Iris data with class labels (top left). The row-wise IQ values for the corresponding heat-map with the 3$\sigma$ vertical line (top right). Plot of the PCA scores for the iris data. Samples denoted with an X, are the samples with row-wise IQ values above the 3$\sigma$ line (bottom).}
	\end{figure}
	
	\subsection{Obsidian Data Pattern Recognition}
	This dataset contains X-ray fluorescence measurements made on 12 obsidian artifacts and 63 obsidian samples that were collected from four locations in Northern California. The four source locations were from quarries located near: Mount Konocti, Borax Lake, Glass Mountain (Napa County), and Anadel (Sonoma County). We used this dataset demonstrate that the sample-sample associations afforded by uncharted forest analysis could be used to establish the likely origin of the artifacts from measured quantities of the following trace elements: iron, titanium, barium, calcium, potassium, manganese, rubidium, strontium, yttrium, and zirconium \cite{Obsidian}. This dataset has previously been demonstrated as a test-bed for unsupervised and supervised learning methods in chemometrics \cite{kowalski}.
	
	The unknown samples were assigned source labels based on an uncharted forest model made with 200 trees and a tree depth of 2, as well as a nearest neighbor classification (k = 1) after PCA projection onto the first two components of the unknown samples The assignments of the artifact samples to their predicted source classes can be seen in Figure 3. The PCA was created from standard normal variate-scaled obsidian source data that had the X-ray fluorescence measurements of iron, strontium, and manganese removed due to correlation with other variables. The uncharted forest analysis assignments were made from the row-wise IQ values of the rows in the probabalistic affinity matrix, which corresponded to the twelve unknown samples via a maximum voting scheme. The voting scheme was performed to remove analyst bias and to automate the assignment of samples to suspected classes, but the results were equivalent to our visual assessment. 
	
	\begin{figure}[H]
		\centering
		\includegraphics[width=0.95\linewidth]{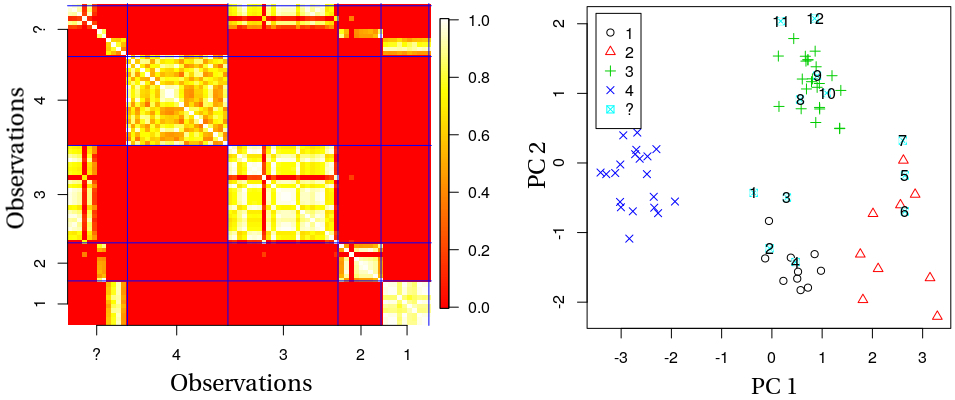}
		
		\caption{The probability affinity matrix (left) and the PCA scores plot obtained from the first two principal components of the standard normal variate-scaled ARCH data (right). The unknown artifact samples are denoted with a `?', and the four source classes of the obsidian data are labeled as follows: Mount Konocti (1), Borax Lake (2), Glass Mountain (3), and Anadel (4). }
	\end{figure}
	
	The source assignments of the artifacts from the uncharted forest analysis with the voting scheme differed from those made from the nearest neighbor (1-NN) PCA classification model by only one sample. For a sense of comparison, an ordinary random forest classification model with 200 trees was also trained on the source data. Again, the assignments made by the maximum vote uncharted forest analysis differed by only one sample from those of the random forest model. However, the classification produced by the nearest neighbor classifier differed from the assignments made by the random forest model by two samples. None of the models assigned any of the artifact samples to the Anadel source class. This result is consistent with a classification analysis performed by a linear learning machine \cite{kowalski}.
	
	Similar assignments between unknown samples and assigned source classes were obtained by the uncharted forest method and the two supervised classification models. This finding suggested that the unsupervised partitioning method used in the uncharted forest method with a maximum voting scheme can offer sample-sample associations that are comparable to that found using supervised methods.

	\subsection{Nan'ao One Shipwreck Data EDA}
	This dataset was selected because it was relatively high in dimension, and because it possessed complex class labels and sample membership. Twelve porcelain samples were collected from a shipwreck and compared, by using NAA measurements of 27 elements, with samples from several nearby kilns. The data were scaled by applying a base ten logarithm prior to analysis as suggested by the authors \cite{shipwreck}. The goal of the study was to determine which kilns were most likely used to create the blue and white porcelain artifacts that accounted for 95\% of the samples found on the wreck \cite{shipwreck}.
	
	An uncharted forest analysis using 30 trees and a tree depth of 2 was used to investigate the source and artifact data, as shown in Figure 4. It was found that two of the artifact samples had high associations with source samples obtained from the Jingdezhen kiln, and no associations with other source samples. This result implied that those samples could be assigned to the Jingdezhen kiln. Zhu et al.'s PCA analysis agreed with the assignment of those two samples to the Jingdezhen kiln \cite{shipwreck}. 
	
	\begin{figure}[H]
		\centering
		\includegraphics[width=0.65\linewidth]{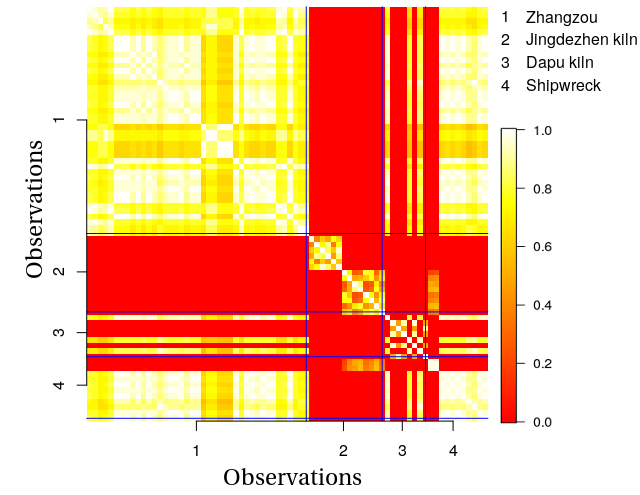}
		
		\caption{Probability affinity matrix of the three source classes and the shipwreck samples. }
	\end{figure}
	
	The probabilistic affinity matrix also indicated that samples from the Jingdezhen kiln may be composed of two subclasses. Our PCA analysis did not readily show that the Jingdezhen kiln could be represented as two subclasses. It was found that the scores of the subclassess that were identified by uncharted forest analysis were linearly separable after projection from the fourth and fifth principal components. Manual assessment of the 729 pairs plots showed that the two subclasses identified by uncharted forest were linearly separable in the Lutetium measurements. The base-10 logarithmically scaled Lutetium variable was was used in 16\% of all decision boundaries made from the random selections of variables and was the third most used variable for constructing decision boundaries in all of the uncharted forest trees. In every instance the decision boundaries were created at -0.569, and this boundary was never imposed at a root node. A plot of normalized kernel density estimates of Lutetium, displayed in Figure 5, showed that the samples from these two possible subclasses were well-separated by the Lutetium decision boundary. This information was readily accessible by uncharted forest analysis, whereas, by the other EDA methods we did not know that this division of samples in the Jingdezhen class may be of interest, or that it existed. The reason why the Jingdezhen class has two well-separated distributions of Lu measurements is not known, but this finding could be of value for future investigations or to provide information about appropriate sample selection that accounts for the sources of variation, as is necessary in the training of classification models. 
	
	\begin{figure}[H]
		\centering
		\includegraphics[width=0.75\linewidth]{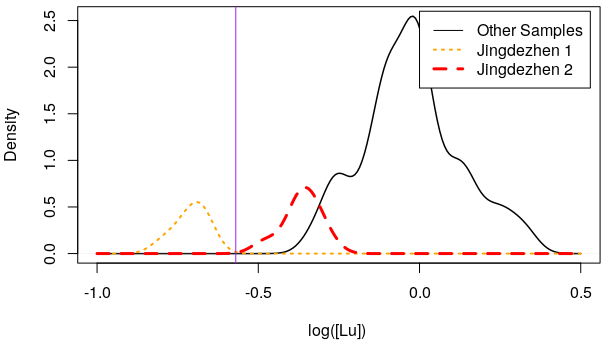}
		
		\caption{ Kernel density estimates for the base-10 logarithmically scaled Lutetium concentrations in the speculated Jingdezhen 1 and 2 subclasses as well as in the other samples. The kernel density estimates were created using a bandwidth of 0.04 and were normalized relative to the kernel estimates of the entirety of the data. The decision boundary created by the uncharted forest algorithm is shown as a vertical line. }
	\end{figure}
	
	Uncharted forest analysis revealed that the remaining artifact samples were found to be most associated with those from the Zhangzou province. However, there was notable interaction between samples from the Zhangzou province and three source samples from the Dapu kiln. Similarly, Zhu et al. found that the remaining samples appeared to be from the Zhangzou province, and that several samples were outside of the 90\% confidence bounds of their principal component analysis \cite{shipwreck}. 
	
	To make the partitions of uncharted forest more specific to the samples believed to be in the Zhangzou province, both the source and artifact samples attributed to the Jingdezhen kiln were removed. The source samples collected from the Dapu kiln were also removed to focus the partitioning only on the kilns in the Zhangzou province. The goal of this analysis was to assess the samples obtained from specific kilns within the Zhangzou province as they related to the shipwreck, rather than assuming the Zhangzou province was a homogeneous collection of source kilns. The probabilistic affinity matrix, shown in Figure 6, produced by uncharted forest analysis with 75 trees and a tree depth of 4 displayed no associations (IQ = 0) between the samples from the shipwreck and the Hauzilou and Tiankeng kilns. This finding suggests that both kilns were unlikely sources for the artifacts, a finding that was not observed in the original study\cite{shipwreck}.
	
	\begin{figure}[H]
		\centering
		\includegraphics[width=0.65\linewidth]{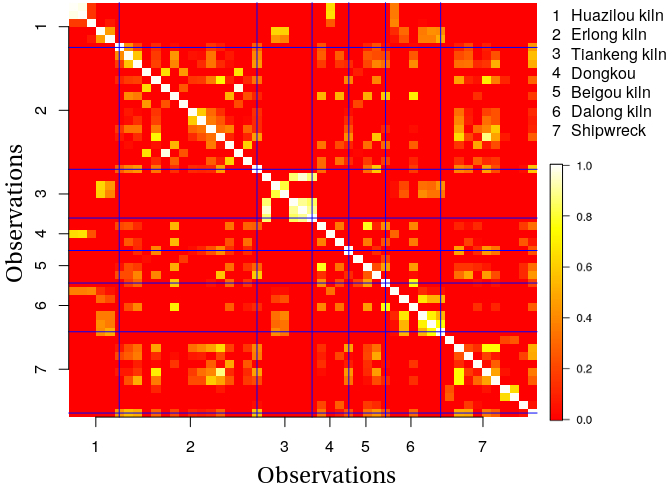}
		
		\caption{The probabilistic affinity matrix after the fragmentation of the Zhangzou province into its respective source kilns and the removal of the two artifact samples associated with the Jingdezhen kiln, and the sources from the Jingdezhen kiln and the Dapu kiln. }
	\end{figure}

	Although the additional analysis did not clearly display which individual kilns in the Zhangzou province were most likely associated with the artifact samples, it did discount two unlikely locations. In a provenance study it is also of interest to know where samples likely did not originate from, because that information can be used for the design of future studies. 
	
	\subsection{Cryptocrystalline Sillicate Data EDA}
	Hoard et al. investigated the geological source of several cryptocrystalline sillicate artifacts that appeared identical by eye and by microscope \cite{crypto}. Artifacts were collected in Eckles, Kansas and at three sites in Nebraska (Signal Butte I/II/III). Eleven trace elements in the artifacts were quantified by NAA and compared to source materials collected from Flattop Butte, Gering formation, Nelson Butte, Table Mountain, and from West Horse Creek. 
	
	The Gering formation TIQ was calculated to be $\ll$ 5\% by an uncharted forest model that had 100 trees and a tree depth of 4. This result suggested that the Gering formation was not commonly partitioned with samples outside of its assigned class label, in agreement with the authors' key finding that the Gering formation was not a likely source for the artifacts\cite{crypto}. This finding is a clear example of the identification of an uninformative class by uncharted forest analysis. This finding was corroborated from the scores obtained by a PCA constructed using only the source classes. However, the PCA scores plots were uninformative with regards to obtaining other information Hoard et al. had described using canonical discriminant analysis (CDA), or that observed via the probabilistic affinity matrix obtained from uncharted forest analysis. The score plots of the first four principal components showed nondistinct overlap of almost all of the samples belonging to the classes, even after the CDA-separable Gering formation class was removed. 
	
	The probabilistic affinity matrix obtained by uncharted forest analysis using 100 trees and a tree depth of 4 with only the Eckles artifact class is displayed in Figure 7. This Figure clearly shows that the most likely source for the samples collected at the Eckles site was the Flattop Butte. This finding by uncharted forest concurred with the CDA analysis performed by Hoard et al \cite{crypto}. 
	
	
	
	\begin{figure}[H]
		\centering
		\includegraphics[width=0.70\linewidth]{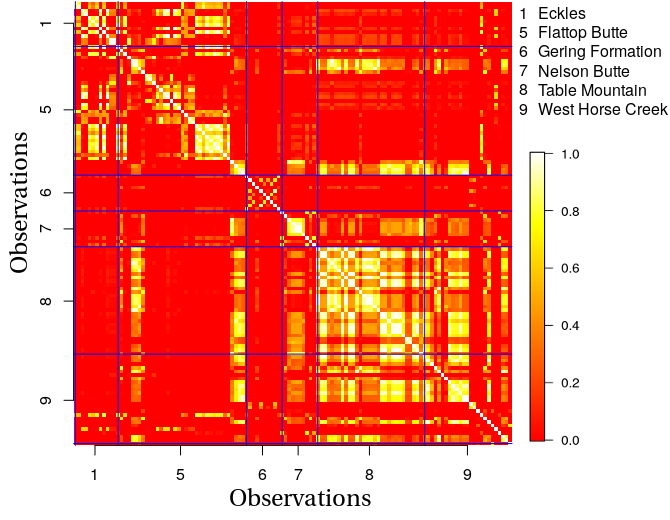}
		
		\caption{The probabilistic affinity matrix from uncharted forest clustering of the cryptocrystalline sillicate data with only the Eckles Kansas artifact and the source classes included. Class boundaries are shown by the overlays of vertical and horizontal lines. }
	\end{figure}
	
	However, unlike the CDA analysis performed by Hoard et al. or the PCA analysis explored in this study, an uncharted forest analysis displayed class heterogeneity. A very clear example of a class with heterogeneous structure, shown in Figure 8, was observed via uncharted forest analysis using 100 trees and a tree depth of 4, by comparing associations between artifacts from the Signal Butte III site and the source locations. Seven of the twelve samples collected from the Signal Butte III site were most frequently partitioned with two samples collected from Table Mountain, not the Signal Butte III samples. Those two samples that were collected from Table Mountain did not appear to be associated with the rest of the data that was labeled as Table Mountain. This instance of class heterogeneity may be indicative of a missing source class, the introduction of a sampling error, or of another anomaly in the Table Mountain data. Hoard et al. was unable to determine the likely origin for several of the samples that were collected from the Signal Butte III site; CDA indicated that the samples from Signal Butte III were tightly grouped \cite{crypto}. Uncharted forest analysis, on the other hand, indicated that the samples may actually be from two groups, one not well-represented in the NAA data, and the other most like the samples obtained from Table Mountain.
	
	\begin{figure}[H]
		\centering
		\includegraphics[width=0.70\linewidth]{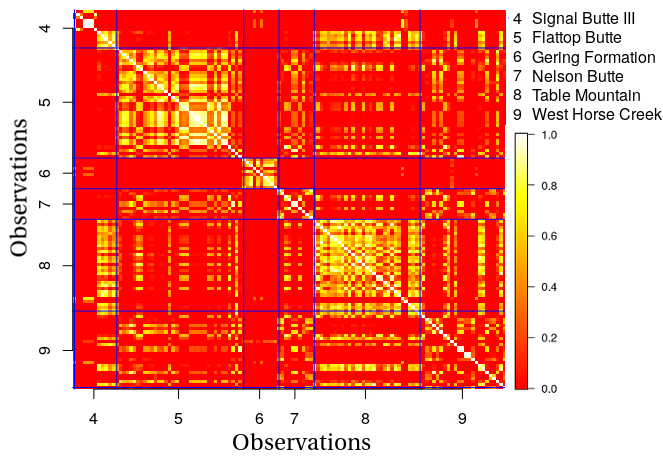}
		
		\caption{The probabilistic affinity matrix from uncharted forest clustering of the cryptocrystalline sillicate data, with overlaid class boundaries. Heterogeneity within the first 7 samples collected from the Signal Butte III site can be observed in that they only associate with two samples collected from Table Mountain. These samples do not associate with the rest of the data from Table Mountain. }
	\end{figure}

	\subsection{Geographical Butter Data Clustering Analysis}
	The butter dataset consists of 78 butter samples collected from 17 countries. The goal of this study was to assess the viability of determining the origin of butter samples from their chemical signatures \cite{butter}. Butter samples were analyzed by inductively coupled plasma mass spectrometry (ICP-MS) and isotope-ratio mass spectrometry using isotopic ratios of $^{13}$C, $^{15}$N, $^{18}$O, and $^{34}$S \cite{butter}. Of the 17 countries included in the data, 8 were represented by two samples or less; these data were removed. A total of 65 samples were retained for analysis, of which 44.6\% were of German origin. A significant proportion of the measurements of $^{34}$S in butter proteins were lower than the detection limit \cite{butter}. To make use of the information in this variable, the samples with $^{34}$S values lower than the detection limit were imputed with zero concentrations. 
	
	In this study, the butter dataset was used to compare the variance partitions made by the uncharted forest algorithm with two variance-based clustering algorithms. K-means clustering and hierarchical clustering with a Ward's criterion were applied to the butter data, using a range of clusters (2-7), with 15 replicates. An uncharted forest analysis with 200 trees and a tree depth of 5 was used to assess each replicate of the clustering methods. K-means clustering optimizes the location of cluster partitions by minimizing within-cluster variance in an iterative fashion \cite{kmeans}. Ward's clustering was employed with a loss function that minimizes within-cluster variance by forming clusters using an agglomerative approach\cite{ward}. The goal of this study was to assess how the TIQ and TSAQ metrics compared to within- and between-cluster variance metrics, respectively. 
	
	Confidence intervals of the correlation coefficients between uncharted forest metrics and the empirical cluster variances for both clustering methods were created by Fisher transforming the correlation coefficients, calculating the confidence intervals as usual, and transforming back \cite{Fisher}. All intervals reported here were computed at 99\% confidence. TSAQ and within-cluster variance were found to be strongly correlated: K-means clustering had a correlation of 0.90 $\pm$ 0.12 with TSAQ, and Ward's clustering had a correlation of 0.94 $\pm$ 0.05. This result strongly suggested that the relationship between TSAQ and the within-cluster variance metric were well approximated by a linear relationship. The intervals for the correlation coefficients between TIQ and between-cluster variances were -0.50 $\pm$ 0.07 for K-means clustering and -0.60 $\pm$ 0.05 for Ward's clustering. Overall, the correlation between TIQ and between-cluster variance of either clustering method was moderate; this finding indicated that TIQ provides information that is only somewhat similar in direction to between-cluster variance.
	
	The negative correlation between TIQ and between cluster variances was expected, because as the number of clusters increases, the cluster centroid distances also increase, and so do the between cluster variances. However, as cluster distances increase, the associations found by uncharted forest analysis between the samples inside of those clusters would be expected to become less prevalent, because the clusters are becoming more specific to the groups in the data; thus, TIQ values would be expected to decrease. The magnitude of the correlations were not near-perfect (-1.0) because TIQ and TSAQ metrics result from randomly created partitions and the uncharted forest algorithm does not partition data the same way that either K-means or Ward clustering does. However, the fact that the correlation was correct in sign suggests that the partitioning mechanism is overall variance-based, as intended.

	\section{Conclusion}
	We have presented a novel adaptation to the random forest paradigm which extends it to unsupervised exploratory data analysis. The method allows for high- and low-dimensional data to be represented as a matrix where every entry is a sampled probability-like value that represents the likelihood that a given sample (row) resides in the same terminal node as the other samples (column). This exploratory data analysis tool and its associated metrics are shown to be a informative aid for the analysis of classification and provenance data. Uncharted forest analysis was able to provide information related to uninformative class labels, class heterogeneity, and sample-wise relationships. 
	
	The uncharted forest analysis method using variance partitions was explored by comparing its measurable results with both between- and within- cluster variances of two known variance based clustering algorithms. It was shown that the uncharted forest algorithm can provide metrics (TIQ, TSAQ) that are correlated to global clustering statistics. However, the true utility of the method is found in the visualization of sample-wise information, and how it relates to cluster/class-wise associations. As a proof-of-concept though, when sample-wise variance partitions were used in uncharted forest analyses, the probabilistic affinity matrix generated metrics that were correlated to rigorously defined statistical measures of spread for clustered data. 
	
	Very similar findings to the classification based provenance studies performed by Zhu et al. and Hoard et al. were obtained from exploratory uncharted forest analysis. This is of interest because the uncharted forest algorithm is unsupervised, but with an overlay of class labels, it was possible to visually infer class labels for unknown samples in high-dimensional, multiclass datasets. Most importantly, uncharted forest analysis was able to display information about the data that was not readily apparent from the classification methods used in the aforementioned studies and other exploratory data analysis methods. The uncharted forest method was shown to be an EDA technique that can provide information that is relevant to the domains of exploratory data analysis, clustering, and classification.
	
	\section{Acknowledgments}
	This work was supported by the United States National Science Foundation grant 1506853.
	
	\section{References}
	
	
	
	
	\bibliographystyle{model1-num-names}
	\bibliography{sample.bib}
	
	

\end{document}